\pdfoutput=1

\documentclass[11pt]{article}

\usepackage[]{acl}

\usepackage{multirow}
\usepackage{graphicx}  
\usepackage{natbib}    
\usepackage{colortbl}
\usepackage{adjustbox}

\usepackage{times}
\usepackage{latexsym}

\usepackage[T1]{fontenc}

\usepackage[utf8]{inputenc}

\usepackage{microtype}

\title{Dialogue Policies for Confusion Mitigation in Situated HRI}

\author{ Na Li, Robert Ross \\
  School of Computer Science \\
  Technological University Dublin  \\ 
  \{na.li, robert.ross \}@tudublin.ie \\
}

\begin{document}
\maketitle
\begin{abstract}
Confusion is a mental state triggered by cognitive disequilibrium that can occur in many types of task-oriented interaction, including Human-Robot Interaction (HRI). People may become confused while interacting with robots due to communicative or even task-centred challenges. To build a smooth and engaging HRI, it is insufficient for an agent to simply detect confusion; instead, the system should aim to mitigate the situation. In light of this, in this paper, we present our approach to a linguistic design of dialogue policies to build a dialogue framework to alleviate interlocutor confusion. We also outline our sketch and discuss challenges with respect to its operationalisation.

\end{abstract}

\section{Introduction}
Confusion is a type of dynamic mental state, which can not only lead to negative conditions, \textit{i.e.}, frustration, boredom or subsequent disengagement in a task or a conversation, but can also be associated with positive conditions as a user seeks to overcome initial confusion \citep{DMelloConfusionlearning2014,li2021detecting}. In mainstream human-computer interaction (HCI) studies, a number of studies have investigated confusion state effects in the context of online learning and driver assistance \citep{eegconfusionlevel,Grafsgaard2011,Zhou2019}. One prominent model of confusion is from \citet{Lodge2018} who pointed to a zone of optimal confusion (ZOC) which is productive confusion, where learners are self-motivated to overcome their confusion state; but also pointed to a zone of sub-optimal confusion (ZOSOC) where learners could not resolve the disequilibrium which in turn leads to confusion persisting such that the confusion becomes unproductive. Similarly, \citet{DMelloConfusionlearning2014} described three bi-directional transitions, \textit{i.e.} confusion-engagement, confusion-frustration and frustration-boredom transitions to explain confusion dynamics. 
Finally, \citet{Arguel2015} presented two thresholds ($T\_a$ and $T\_b$) bounding levels of confusion potential in learning. Between the two thresholds is the confusion stage, and if the level of confusion is less than $T\_a$, then the learners should be fully engaged, whereas if the confusion level is over $T\_b$ the confusion is not mitigated leading to learners becoming bored.

However, little work has focused on confusion detection and modelling in general conversational interactions or human-robot interaction (HRI). Given this gap, in our research, we aim to detect,  model, and in time mitigate confusion states (\textit{i.e.} productive confusion, unproductive confusion). For this work, we focus on four confusion induction types, \textit{i.e.}, complex information, contradictory information, insufficient information, and false feedback \citep{LEHMAN2012,Lehman2013WhoBF,Silvia2010ConfusionAI}. 

Although our work to date has focused on confusion \cite{li2021detecting,naliembodiment2022}, modelling and detection, it is also essential that the dialogue agent is capable of mitigating user confusion and helping participants reengage in the ongoing task-oriented interaction. 
Our model is based on seven dialogue act types that are used to implement strategies for confusion mitigation. In light of this need, in the paper, we sketch out our initial approach to design a dialogue policy for task-oriented interaction that can be used to mitigate users confusion states if identified. The model consists of a general dialogue policy and two specific policies for different confusion induction situations. While HRI includes verbal and nonverbal interactions \cite{hribook2020}, in this initial work, our outline dialogue policies are restricted to linguistic interactions.

\section{Act and Policy Outline}

As the basis of the policy combining the specific case study of confusion mitigation, we first outline a sort of dialogue act types corresponding to a general dialogue policy, and then two sub-policies for two confusion states mitigation are produced. Therefore, we start by introducing the following seven key dialogue act types and highlight their relevance to the mitigation as follows:

\begin{enumerate}
\item \textbf{Restatement}: The agent repeats the information or question.
\item \textbf{Feedback request}: The agent asks for the participant's feedback and response.
\item \textbf{Information extension}: The agent provides more information to expand on the information or question already raised.
\item \textbf{Information supplement}: The agent provides comprehensive information or questions in different ways for participants to quickly understand easily. 
\item \textbf{Response correction}: The agent provides the appropriate response in order to avoid confusion states on the participant.
\item \textbf{Confirmation}: The agent admits that the information or question has one or more issues leading to the participant being confused. 
\item \textbf{Subject change}: The agent changes straightforward questions or other topics.
\end{enumerate}

\begin{figure}[ht]
    \centering
	\includegraphics[width=0.4\textwidth,height=10cm]{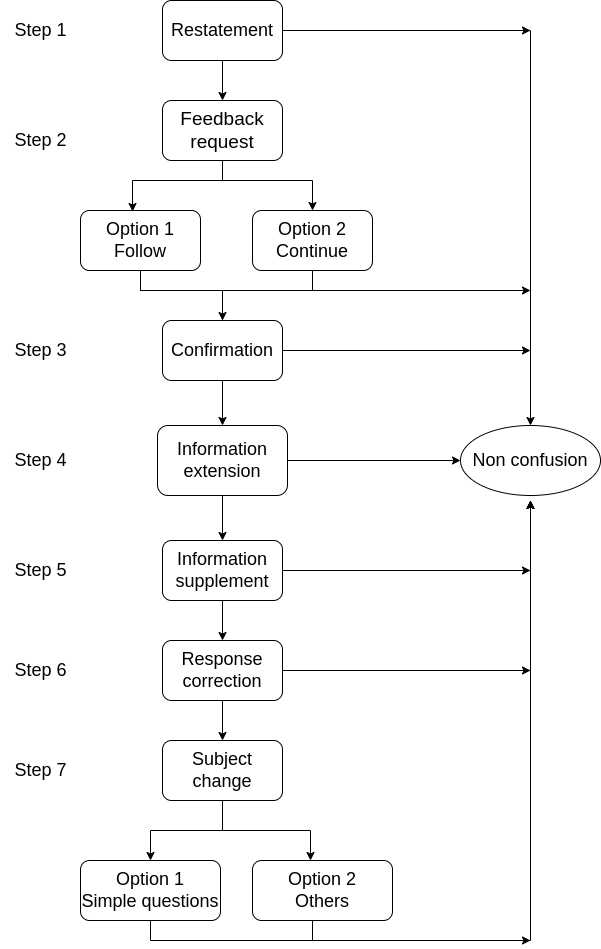}
	\caption{\label{fig:policyprocess} General policy process of confusion mitigation}
\end{figure}

We applied the seven types of dialogue act to first design a general dialogue policy based on a number of communicative rules (see Table 1). Figure \ref{fig:policyprocess} illustrates the operating dialogue policy as a control flow process, with each step corresponding to one of the detailed elements of the outline rules in Table 1. In this control flow policy, each step makes it possible to help users who are confused transfer to a non-confusion state. If after any one step, the user's confusion still cannot be mitigated, then the agent will move to the next step.

Based on this general framework policy, we have developed a set of sub-policies to apply in the specific cases of productive and unproductive confusion in the case of the four confusion induction types mentioned earlier. The first of these dialogue sub-policies (see Table 2) includes the dialogue act types and corresponding communication rules to reduce productive confusion according to the induction of a specific confusion method. The second sub-policy (see Table 3) addresses the case where the participant has reached an unproductive confusion state, where they may be frustrated or even want to drop the conversation. Therefore, this sub-policy helps the participant reengages in interacting with the agent from their unproductive confusion state. The three detail policies in Table 1, Table 2 and Table 3 are mentioned early, \textit{i.e.} general dialogue policy, and two sub-policies for mitigating productive and unproductive confusion are attached to GitHub \footnote{Table 1, 2, 3: https://github.com/lindalibjchn/dialoguepolicy.git}.


\section{Discussion \& Outlook}

Although this short paper simply provides a sketch of our approach, we are building on this sketch to implement a physical test for those policies based on a wizard-of-oz study \cite{Riek2012WizardOO} using physical situated robots integrating our existing platform. We expect that this work can drive a true formalisation and evaluation of these policies. Therefore, our goal is to fully operationalise this policy, but this, of course, is non-trivial. While we could aim to formalise this model through an appropriate formalisation, such as type theory with records (TTR), a Machine Learning (ML) driven approach would be more suitable for a robust system construction. Ultimately, our goal is to develop a hybrid policy that can have general structures to accommodate the user state, but is driven by a probabilistic framework.


\section*{Acknowledgements}
This publication has emanated from research conducted with the financial
support of Science Foundation Ireland under Grant number 18/CRT/6183. For the purpose
of Open Access, the author has applied a CC BY public copyright licence to any
Author Accepted Manuscript version arising from this submission.

\bibliography{nali}

\appendix

\end{document}